%% file: main.tex
\documentclass[10pt,twocolumn,letterpaper]{article} 

\usepackage{avss}
\usepackage{times}
\usepackage{epsfig}
\usepackage{graphicx}
\usepackage{amsmath}
\usepackage{amssymb}
\usepackage{makecell}
\usepackage{enumitem}
\usepackage{multirow}
\usepackage[export]{adjustbox}
\usepackage{multicol,lipsum}
\usepackage{marvosym}  

\usepackage{graphicx}
\usepackage{booktabs}

\input{preamble}

\usepackage[pagebackref=true,breaklinks=true,letterpaper=true,colorlinks,bookmarks=false]{hyperref}

\avssfinalcopy 

\def\avssPaperID{112} 

\ifavssfinal\fi

\begin{document}

\title{\emph{Dare to Plagiarize?} Plagiarized Painting Recognition and Retrieval}

\author{Sophie Zhou \\
Cheyenne Mountain High School\\
{\tt\small zhoul.sophie1204@gmail.com}
\and
Shu Kong$^{\text{\Letter}}$\\
University of Macau\\
{\tt\small skong@um.edu.mo}
}

\maketitle


\begin{abstract}

Art plagiarism detection plays a crucial role in protecting artists' copyrights and intellectual property, yet it remains a challenging problem in forensic analysis. In this paper, we address the task of recognizing plagiarized paintings and explaining the detected plagarisms by retrieving visually similar authentic artworks. To support this study, we construct a dataset by collecting painting photos and synthesizing plagiarized versions using generative AI, tailored to specific artists' styles. We first establish a baseline approach using off-the-shelf features from the visual foundation model DINOv2 to retrieve the most similar images in the database and classify plagiarism based on a similarity threshold. Surprisingly, this non-learned method achieves a high recognition accuracy of 97.2\% but suffers from low retrieval precision 29.0\% average precision (AP). To improve retrieval quality, we finetune DINOv2 with a metric learning loss using positive and negative sample pairs sampled in the database. The finetuned model greatly improves retrieval performance by 12\% AP over the baseline, though it unexpectedly results in a lower recognition accuracy (92.7\%). We conclude with insightful discussions and outline directions for future research.


\end{abstract}


\section{Introduction}
\label{sec:intro}

The mass production of AI-generated art presents a growing thread to the preservation of human artistic expression and the authentication of original works. As the art world becomes increasingly digitized, concerns about uncredited and uncompensated plagiarism of artists' styles are intensifying. 
For example, Vincent van Gogh, renowned for his distinctive brushwork and expressive style, is especially vulnerable to such exploitation. In an era where digital reproductions and deepfakes are becoming widespread~\cite{nguyen2019deepfake}, the need to distinguish genuine artistic intent from automated imitations is more urgent than ever.

Traditional methods of art authentication rely heavily on expert analysis, provenance documentation, and physical examination of the artwork~\cite{stuart2012forensic}. However, these methods are often time-consuming, costly, and somewhat inconclusive due to subjectivity, especially when faced with high-quality forgeries. Techniques of computer vision and deep learning offer promising alternatives~\cite{galambos2013forensic} to facilitate recognition of plagiarized artworks and explain how they plagiarize.
In this work, we study plagiarized painting identification and retrieval, with the latter helping explain how query paintings plagiarize existing artworks.

To support this study,
we construct a database consisting of downloaded paintings of van Gogh and other artists. 
To simulate plagiarized paintings, we focus on van Gogh's artwork~\cite{woodward2009art} and use generative AI (ChatGPT in this work) to synthesize plagiarized versions.
With the database,
we first evaluate a non-learned method that first extracts features of both the query painting and all photos in the database using the visual foundation model DINOv2~\cite{oquab2023dinov2},
then computes cosine similarities between the query and the photos in database, and finally returns top-ranked similar photos.
It thresholds the similarity score to classify the query as either ``\emph{authentic}'' or ``\emph{plagiarized}''.
This non-learned method achieves 97.2\% classification accuracy but only 29.0\% precision in retrieval.
To improve over this baseline, we finetune DINOv2 using a metric learning loss over positive and negative paired examples sampled from the database.
The finetuned model significantly increases retrieval precision to 41.2\% but yields lower classification accuracy (92.7\%).
We hypothesize that the AI-generated plagiarized photos are more easily recognized by the pretrained visual foundation model DINOv2, which, however, can struggle to do so if being finetuned (as finetuning destroys the generality of pretrained features~\cite{kumar2022finetuning}).
Nevertheless, we provide more insightful discussions and outline future research directions.

We make three contributions:
\begin{enumerate}[leftmargin=15pt, topsep=0pt, itemsep=5pt,parsep=-2pt]
    \item We introduce a dataset consisting of authentic and AI-generated plagiarized paintings to support the study.
    \item We develop a non-learned baseline and a learning-based method that demonstrate promising results in plagiarism recognition and retrieval.
    \item We provide insightful discussions and outline directions for future research on this topic.
\end{enumerate}

\section{Related Work}
\label{sec:related-work}

\textbf{AI Generated Content.} 
AI has been increasingly used to create visual content, leading to a surge in AI-generated images and even artworks.
Approaches based on generative AI have demonstrated the ability to produce creative output that challenges human-made content~\cite{goodfellow2014generative, radford2018improving}. Recent studies on neural art synthesis have further advanced this capability, showing that AI can emulate complex artistic styles~\cite{elgammal2017can}.
Our work concerns about recognizing the AI-generated plagiarized artworks and explaining how they plagiarize through image retrieval.

\textbf{Foundation Models} have significantly reshaped the landscape of computer vision research~\cite{bommasani2021opportunities}, with DINOv2~\cite{oquab2023dinov2} representing an example in visual foundation models. 
Moreover, foundation models offer features that capture semantic information~\cite{caron2021emerging} and show impressive generalization to diverse downstream tasks.
In this work, we adapt DINOv2 for plagiarized painting recognition.


\textbf{Image Retrieval}
aims to find similar or relevant images from a database given a query image~\cite{zheng2017sift}.
Contemporary methods end-to-end train deep neural networks on a large-scale of training data\cite{shafique2024crisp}, develop techniques to capture local and global features to enhance retrieval performance~\cite{weinzaepfel2022learning}.
Instead of studying image retrieval methods,
we use it to retrieve relevant photos in the database to help explain how the query painting is plagiarized by returning highly similar photos to help investigation.




\section{Dataset and Benchmarking Protocol}


\subsection{Dataset Construction}

We construct our dataset by (1) downloading painting photos of Van Gogh and other artists, and (2) using generative AI (ChatGPT in this work) to synthesize plagiarized painting photos of Van Gogh.
We tag these images with three categories and split them into training, validation, and testing data:
\begin{enumerate}[leftmargin=15pt, topsep=0pt, itemsep=5pt,parsep=-2pt]
    \item \emph{Authentic Van Gogh paintings} are downloaded online and split into 300/100/100 training/validation/testing images.
    
    \item \emph{Other paintings} are downloaded painting photos from other artists such as Monet. We split them into 300/100/100 training/validation/testing images.
    
    \item \emph{Plagiarized images} are generated by prompting ChatGPT  to mimic the style of  uploaded paintings of Van Gogh.
    We obtain into 300/100/100 training/validation/ testing images.
\end{enumerate}
We combine the three sets of training images to develop models,  use the validation images to tune hyperparameters pertaining to model training,
and evaluate models over the testing images.
We resize all images to 224x224 resolution.

\subsection{Evaluation Metrics}
We evaluate models in terms of plagiarism recognition and retrieval.
For recognition metrics, we adopt the widely-used top-1 accuracy measuring the performance of models to classify a testing image as either ``authentic'' or ``plagiarized''.
For retrieval, we use mean Average Precision (mAP~\cite{manning2008introduction}):
\begin{itemize}[leftmargin=15pt, topsep=0pt, itemsep=5pt,parsep=-2pt]
\item If the query (i.e., testing image) is a plagiarized Van Gogh painting, the positive retrieved examples should be Van Gogh's paintings.

\item If the query is an authentic Van Gogh's painting, the positive retrieved examples should be from Van Gogh.

\item If the query is another artist's painting, the positive retrieved examples should be from other's artwork and Van Gogh's are negatives.
\end{itemize}
To compute mAP, for each query, we rank
the examples in the database and computer Average Precision (AP):
\[
AP = \sum_{k=1}^{N} P(k) \Delta R(k),
\]
where $P(k)$ represents the precision at cutoff $k$ (a threshold on similarity score), and $\Delta R(k)$ is the incremental recall. The mAP is the averaged AP over all testing images.



\section{Methodology}

We present two approaches, a non-learned baseline and a learning method that finetunes a  foundation model.

\subsection{A Non-Learned Baseline}


We use the foundation model DINOv2~\cite{oquab2023dinov2} to extract features for the query image and all images in the database.
Over the features, we compute cosine similarity of the query image and images in the database.
We use the similarity score to rank images of the database for retrieval, and threshold them to classify the query image as either ``authentic'' or ``plagiarized''.
We search for the threshold by maximizing classification accuracy on the validation set.
We find that this baseline yields quite high classification accuracy (97.2\% as seen in Table~\ref{tab:benchmark_results}) but low retrieval mAP (29.0\%).
Next, we introduce a learning-based method to improve the retrieval performance.





\begin{figure*}[th]
  \centering
  \begin{minipage}[t]{0.49\linewidth}
    \centering
    \includegraphics[page=1, width=\linewidth]{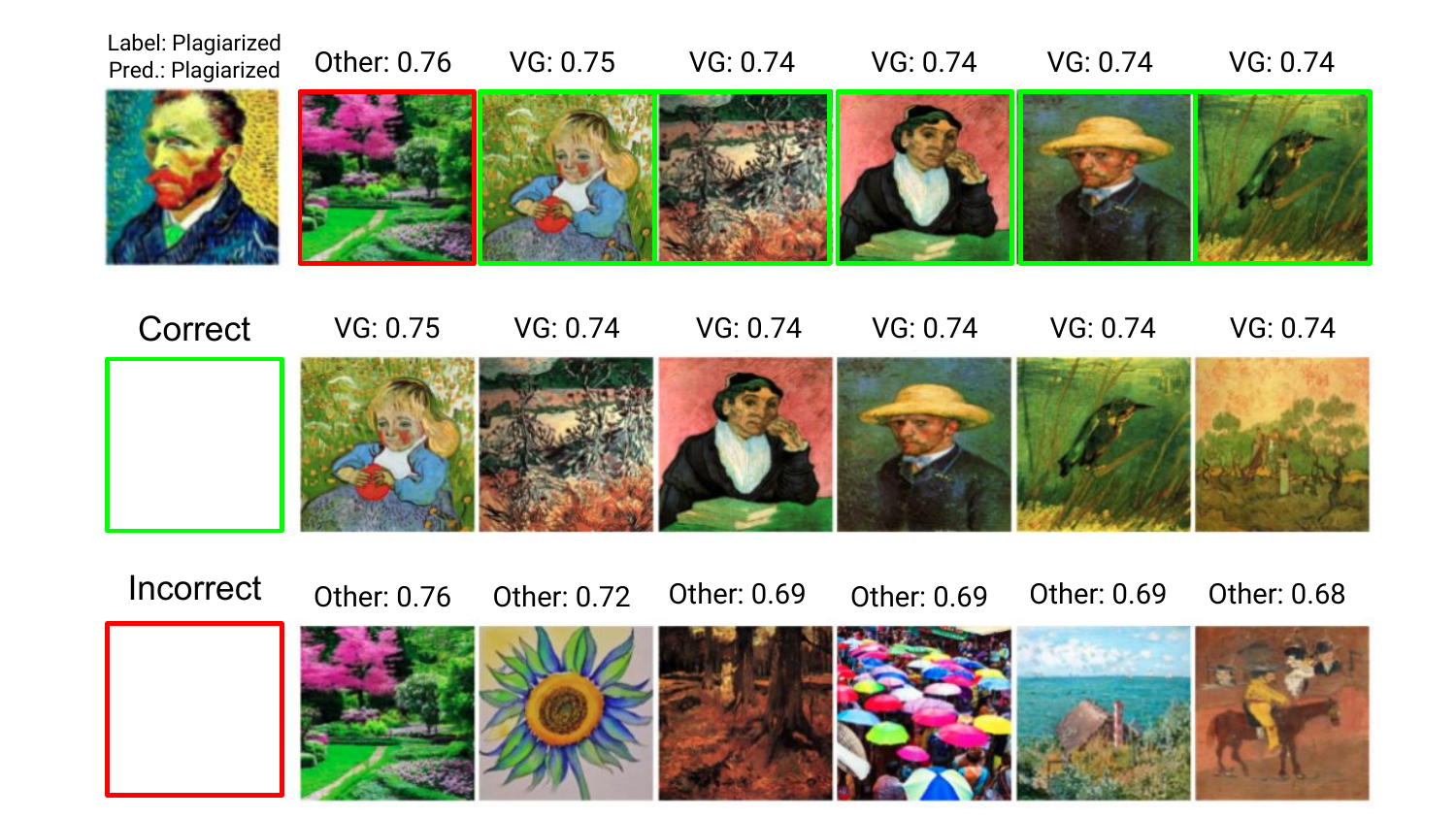}
  \end{minipage}
  \hfill
  \begin{minipage}[t]{0.49\linewidth}
    \centering
    \includegraphics[page=2, width=\linewidth]{figures.pdf}
  \end{minipage}
  \vspace{-2mm}
  \caption{
  We show qualitative results of two queries over a database that excludes AI-generated plagiarized paintings.
  Our model correctly recognizes that these two query examples are not ``authentic'' Van Gogh's paintings.
  We display top-6 retrieved photos with red boxes indicating incorrect examples.  
    For further analysis, we display the top-6 ranked Van Gogh (VG) painting images in the second row and the top-6 photos from other artists.
    This suggests that classification is better done separately rather than using the label of the top-1 ranked example as the predicted label.
    }
\label{fig:name1}
\vspace{-2mm}
\end{figure*}

\begin{figure*}[th]
  \centering
  \begin{minipage}[t]{0.49\linewidth}
    \centering
    \includegraphics[page=3, width=\linewidth]{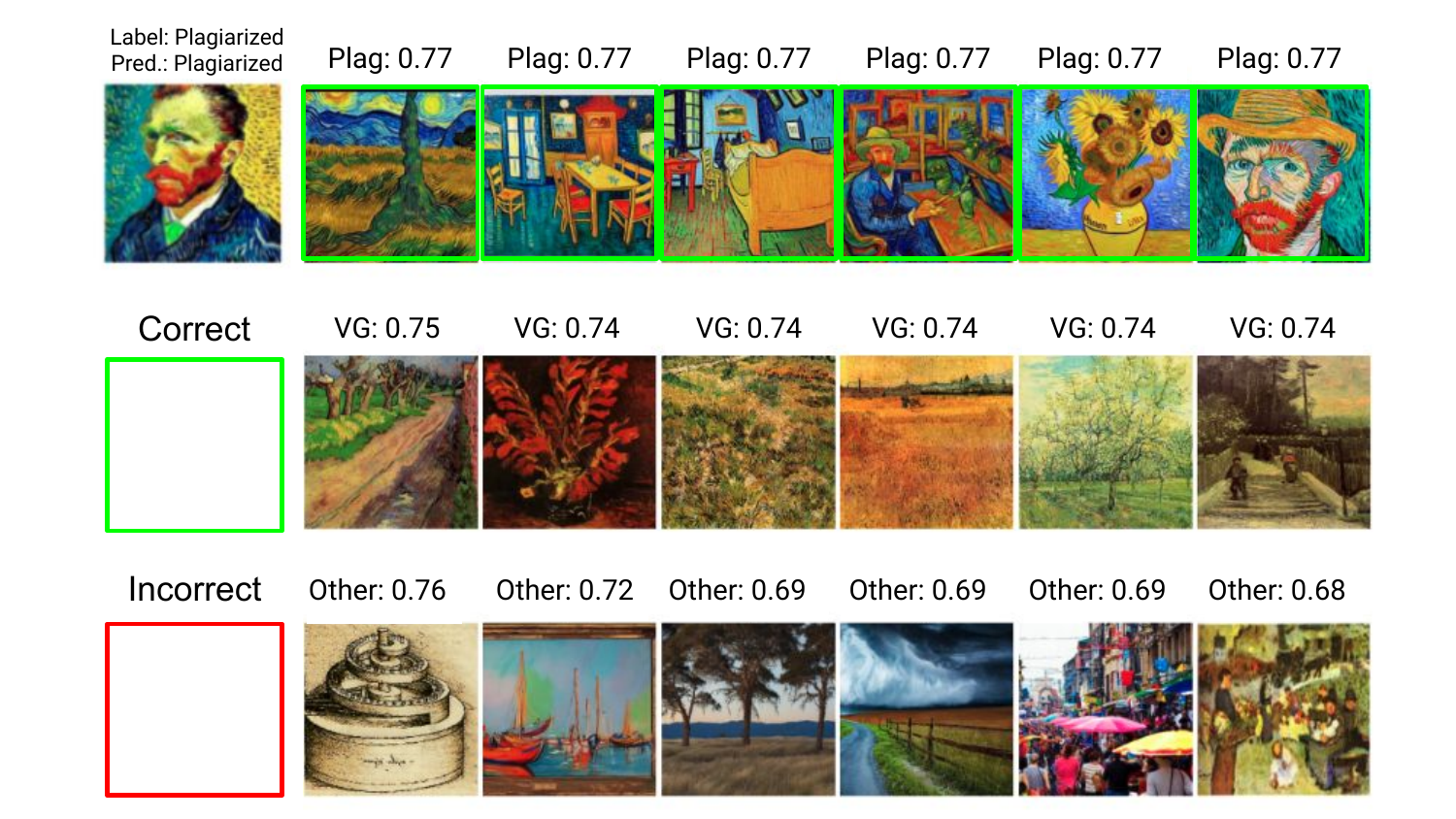}
  \end{minipage}
  \hfill
  \begin{minipage}[t]{0.49\linewidth}
    \centering
    \includegraphics[page=4, width=\linewidth]{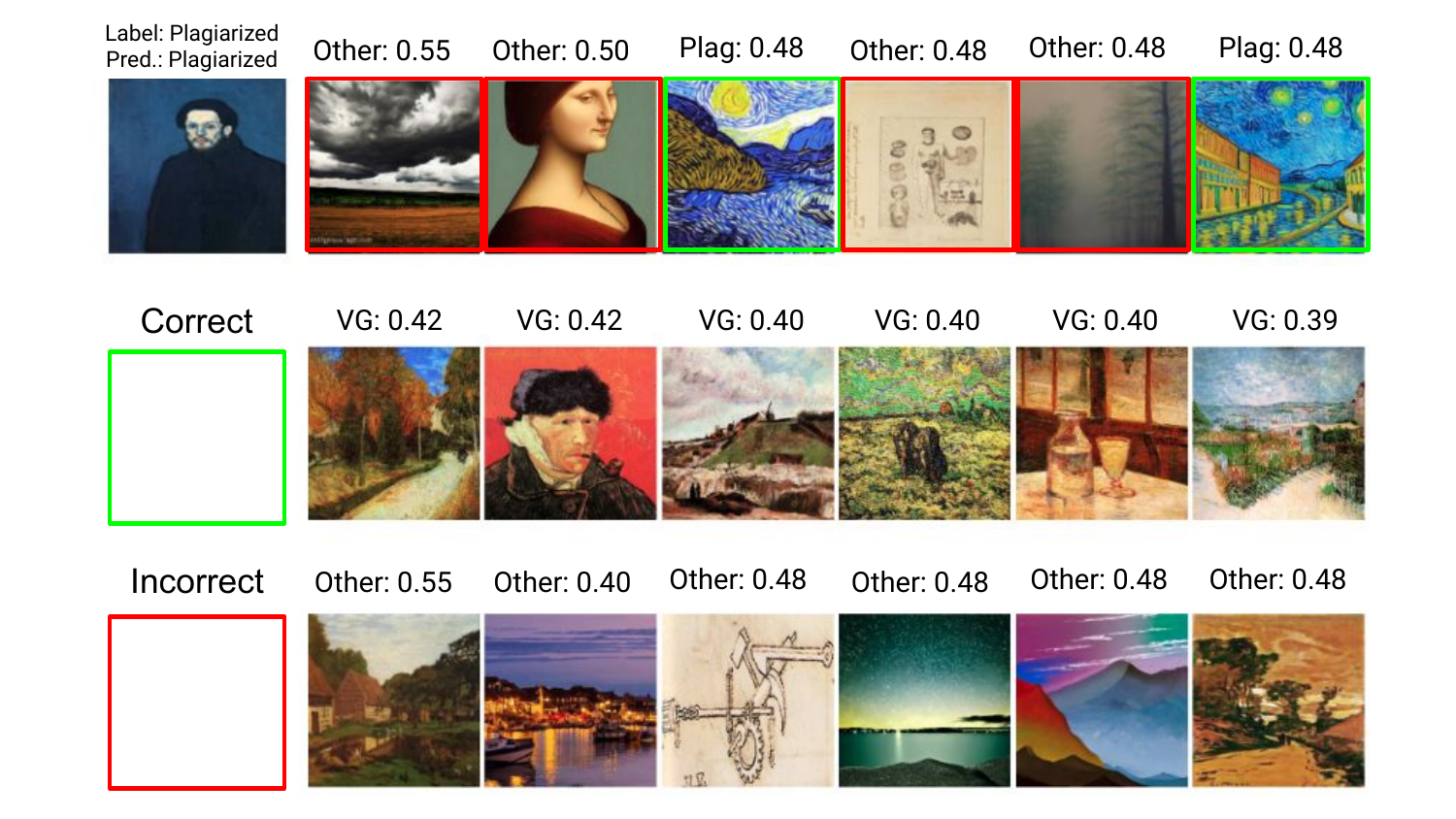}
  \end{minipage}
  \vspace{-2mm}
  \caption{
  We show qualitative results of two queries over a database that contains AI-generated plagiarized paintings.
  Our model correctly recognizes that these two query examples are not ``authentic'' Van Gogh's paintings.
  We display top-6 retrieved photos with red boxes indicating incorrect examples. 
  For further analysis, we display the top-6 ranked Van Gogh (VG) painting images in the second row and the top-6 photos from other artists.
  From the right example, we see that the label of the top-1 ranked example is not that of the query, on which our model correctly recognize its label. This suggests that classification is better done separately rather than using the label of the top-1 ranked example as the predicted label.
}
\label{fig:name2}
\vspace{-2mm}
\end{figure*}

\subsection{Foundation Model Finetuning}

To further improve the retrieval performance, we finetune DINOv2 using a triplet loss~\cite{shen2025solving}.
We sample triplet of training images $(a, p, n)$ where:
\begin{itemize}[leftmargin=15pt, topsep=0pt, itemsep=5pt,parsep=-2pt]
    \item Anchor $a$ as a training image of Van Gogh,
    \item Positive image $p$ as another authentic or plagiarized photo of Van Gogh in the training set,
    \item Negative image $n$ as a training image of other artists.
\end{itemize}
The triplet loss on this triplet is defined as:
\[
\mathcal{L} = \max\big(0,\, d(a, p) - d(a, n) + \alpha\big),
\]
where $d(\cdot,\cdot)$ denotes the Euclidean distance or inverse cosine similarity, and $\alpha$ is the margin (set to 1.0). 
This loss function ensures that the distance between the anchor and positive is at least $\alpha$ less than the distance between the anchor and negative.



{\bf Classifier Learning.}
On top of the finetuned DINOv2, we train a linear classifier on the same training set.
Specifically, we train a Support Vector Machine (SVM)~\cite{cortes1995support} to classify images as either ``authentic'' or ``plagiarized''.

\section{Experiments}

We evaluate our models using both quantitative metrics and qualitative visualizations.
We implement our models on a single A100 GPU.
In our learning-based method, we finetune the top 2 blocks of DINOv2 for 20k iterations, using learning rate 1e-4, batch size 32 and the and the AdamW optimizer.
Below, we discuss our results.


{\bf Quantitative Evaluation.}
Table~\ref{tab:benchmark_results} compares our baseline and learning-based method, and breaks down results w.r.t testing images of Van Gogh (i.e., ``authentic''), other artists, and the ``plagiarized''.
Results demonstrate that the non-learned baseline already achieves high classification accuracy despite low mAP in retrieval.
In contrast, our learning-based method greatly enhances retrieval performance (from 29.0\% to 41.2\% mAP) with accuracy degradation (from 97.2\% to 92.7\%).
This suggests that AI-generated paintings might be easier to identify than one would have thought.
Further,
as finetuned DINOv2 yields degraded classification accuracy despite significant higher mAP in retrieval,
we conjecture that the AI-generated plagiarized photos are more easily recognized by the pretrained visual foundation model DINOv2, which, however, struggle to do so if being finetuned (as finetuning destroys the generality of pretrained features~\cite{kumar2022finetuning}).


{\bf Qualitative Visualizations.}
Fig.~\ref{fig:name1} to \ref{fig:name2} display qualitative results, demonstrating the effectiveness of our model for plagiarism recognition and retrieval.
We refer the reader to the captions for details.

\begin{table}[t]
\centering
\small
\caption{Quantitative comparison of our   \emph{baseline} and \emph{learning} w.r.t classification accuracies and retrieval mAP. 
Perhaps surprisingly, our non-learned method already achieves high accuracy (97.2\%), implying that identifying AI-generated painting photos is a relatively easy task.
In contrast, our finetuned model significantly enhances retrieval but yields degraded classification accuracy.
This suggests that (1) finetuning the foundation model using metric learning indeed facilitates retrieval, but (2) the finetuned model might destroy pretrained features which are generalizable.
}
\label{tab:benchmark_results}
\setlength{\tabcolsep}{2.mm}
\scalebox{0.9}{
\begin{tabular}{lcccccc}
\toprule
method &
Van Gogh & 
Plagiarized& 
Other&
Accuracy&
mAP\\
\midrule

\emph{baseline} & 98.0\% & 96.0\% & 97.5\% & 97.2\% & 29.0\% \\



\emph{learning} & 99.0\% & 81.0\% & 98.0\% & 92.7\% & 41.2\% \\
\bottomrule
\end{tabular}
}
\vspace{-4mm}
\end{table}

\section{Discussion}
Although our results imply that recognizing AI-plagiarized paintings is much easier than one would have imagined,
we point out that sophisticated generative AI techniques could produce more ``authentic'' plagiarisms. 
Furthermore, as we use AI-generated paintings as the plagiarism, we point out that our developed methods might not generalize to real plagiarized paintings, e.g., those that are plagiarized by human artists.
That said, future research should collect a large-scale benchmark dataset to better study plagiarism recognition.

Moreover, as our non-learned method achieves quite high classification accuracy (on plagiarism recognition),
we hypothesize that the pretrained foundation model DINOv2 (and possibly other foundation models too) has learned generic features which capture characteristics of real or fake (AI-generated) images.
It is worth delving into this problem with large-scale datasets.

Lastly, although our work uses image retrieval to help explain how query painting is plagiarized, we note that finer-grained analysis is needed to better understand the plagiarism.
For example, a mature plagiarism detection system should analyze artistic styles, visual elements in the painting, etc.
Achieving these still demands a meticulous dataset.



\section{Conclusion}
We study the problem of plagiarized painting recognition and explaining plagiarism through image retrieval, i.e., demonstrating over which authentic paintings the query is plagiarized.
We present a non-learned baseline that exploits pretrained foundation model DINOv2,
and a learning-based method that finetunes DINOv2 over the database of paintings.
Experimental results demonstrate that recognizing plagiarized paintings is an easy task,
yet retrieving relevant authentic paintings is relatively harder.
Our study suggests the community (1) collect large-scale benchmark datasets to study art plagiarism by including real plagiarized art works as well as AI-generated ones,
(2) develop methods for fine-grained analysis of plagiarized art works.


{\small 
\noindent {\bf Acknowledgements.}
This work was supported by the University of Macau (SRG2023-00044-FST), FDCT (0067/2024/ITP2), and the Institute of Collaborative Innovation.
}

{
\small
\bibliographystyle{ieee}
\bibliography{egbib}
}

\end{document}

%% file: preamble.tex
\def\1{mathbb{1}}

\def\0{{\bf 0}}
\def\1{{\bf 1}}

\makeatletter
\DeclareRobustCommand\onedot{\futurelet\@let@token\@onedot}
\def\@onedot{\ifx\@let@token.\else.\null\fi\xspace}

\usepackage{graphicx}
